# Understanding Learning through the Lens of Dynamical Invariants

*Alex Ushveridze*

*Algostream Consulting, Minneapolis, MN, USA*

## Abstract

*This paper proposes a novel perspective on learning, positing it as the pursuit of dynamical invariants – data combinations that remain constant or exhibit minimal change over time as a system evolves. This concept is underpinned by both informational and physical principles, rooted in the inherent properties of these invariants. Firstly, their stability makes them ideal for memorization and integration into associative networks, forming the basis of our knowledge structures. Secondly, the predictability of these stable invariants makes them valuable sources of usable energy, quantifiable as kTln2 per bit of accurately predicted information. This energy can be harnessed to explore new transformations, rendering learning systems energetically autonomous and increasingly effective. Such systems are driven to continuously seek new data invariants as energy sources. The paper further explores several meta-architectures of autonomous, self-propelled learning agents that utilize predictable information patterns as a source of usable energy.*

## 1. Introduction

The concept of harnessing information as a source of energy is not a novel one, tracing its roots back to the pioneering, albeit flawed, thought experiments of Maxwell and Szilard – namely, the Maxwell's Demon and the Szilard Engine [1,2]. Subsequent works by Rolf Landauer, Charles Bennett, and others [see, for example, 3-6 and references therein] identified errors in the original hypotheses and clarified that the act of observing information, a prerequisite for energy extraction, inherently requires energy investment. This investment effectively nullifies the net energy gain, thereby upholding the second law of thermodynamics. However, this revelation also led to more fascinating conjectures about the interplay between information and energy. Notably, information can indeed serve as an energy source if it is accurately predicted in advance, rather than observed. The key distinction is that prediction can be automated in an energy-neutral manner [7]. The general formula representing this relationship is:

$$[\text{Energy}] \ = kT \ln 2 \ [\text{Correctly Predicted Information}] \qquad (1)$$

Here $T$ represents the absolute temperature and $k$ is Boltzmann's constant. For a detailed explanation and derivation of this formula, refer to [7]. The significance of this correlation lies in its implication that any energy-dependent, open, autonomous system could be incentivized to learn. Learning enhances the predictability of information, thereby increasing the potential to utilize it as an energy resource.

This realization gave rise to the concept of the Autonomous Turing Machine [7] – a theoretical construct that extends the classical Turing Machine to a scenario where the energy required for its operations (including reading and writing on tape, computation, and movement along the tape) is derived directly from the information encoded on the tape. The core hypothesis is that the reliance on correctly predicted information as the sole energy source would inherently 'motivate' this AI agent to enhance its predictive capabilities, thereby naturally evolving towards higher intelligence levels. The term 'motivation' here is used not in its conventional cognitive science context, but rather as a metaphor for certain physical processes within an open system – the agent's body – as it interacts with its environment. The primary objective of this paper is to explore suitable models for studying such systems, with the ultimate goal of shedding light on potential mechanisms through which intelligence might emerge in the universe.

Currently, two primary models are being considered for this purpose. The first, based on the so-called 'thermodynamics of reversible computation' (for details see papers [8,9]), adopts a digital (bit-level) perspective on information. In this model first formulated in [7],, the learning process is akin to a resource-mining operation, driven by thermodynamic (entropic) forces. The second model, which we proposed in our recent paper [10], views computation as an analog (non-digital) process. This approach is grounded in the dynamics of open classical physical systems, specifically those of a resonant nature. Here, thermodynamic principles are not directly applied; instead, they manifest microscopically as dissipative elements within the classical equations of motion.

It is important to recognize that, despite their apparent differences, these two approaches – digital and analog – are fundamentally equivalent and exhibit striking parallels. In this paper, our focus will be primarily on the digital models, though the analog perspectives will also be considered. The preference for digital models stems from their alignment with current AI and ML paradigms regarding information, computation, and algorithms. This alignment renders digital models particularly relevant, especially at a higher conceptual level, for exploring the potential of self-propelled and autonomous intelligence within mainstream AI and ML frameworks. Our aim is to examine potential architectural solutions for the digital variant of the autonomous learning model. By identifying specialized components and delineating their interactions, we hope to gain insights into the possible meta-architecture of AI systems capable of autonomous learning in unfamiliar environments, without the need for dedicated energy sources.

The central thesis of this paper is that despite the apparent diversity and complexity of real-world data, the task of designing fully autonomous, universal learning machines capable of comprehending such data may be more straightforward than initially anticipated. Our confidence in this assertion stems from the observation that, regardless of the complexity of the data flows encountered by these machines, they can invariably be distilled into a universal, concise, and elegant form. This form is characterized by its simplicity, encompassing nothing beyond time and constants. In this context, understanding a system equates to identifying the sequences of data transformations that connect the initial variables, gathered from direct experiments and observations, to this simplified form. Intelligence, therefore, can be aptly defined as the capacity to uncover these transformational chains, linking the raw observed data to sets of dynamical invariants[1].

---

[1] The application of dynamical invariants in machine learning is not a novel concept. There exists a substantial body of literature exploring this topic [11,12], particularly within the realm of Physics-Informed Machine Learning [13-15]. However, these discussions have not typically focused on the role of dynamical invariants as an energy resource for self-propelled learning agents.

The structure of this paper is as follows: Section 1 delves into key aspects and draws parallels between digital and analog models of learning. Section 2 highlights the crucial role of constants in self-propelled learning. In Section 3, we introduce fundamental concepts from dynamical systems theory and apply them to conceptualize an initial design for a self-learning, energy-efficient digital device. Sections 4 and 5 extensively discuss potential meta-architectures for more advanced, universal self-propelled learning machines. Finally, Section 6 addresses the inherent limitations of digital models and explores the challenges that may arise in their practical implementation.

## 2. On Digital and Analog Learning Models

The central formula (1) can be elucidated using the simple model of a one-bit memory unit, exemplified by a cylinder divided into two equal parts by a diaphragm, containing a single atom in one of these compartments. The presence of the atom in the left or right compartment represents binary values 0 and 1, respectively. The read-write operation for this memory is executed using a piston that can be inserted from either side of the cylinder. If the cylinder is in thermodynamic equilibrium with a thermal bath at temperature $T$, and the compartment currently housing the atom is known (perhaps through accurate prediction), then by inserting the piston into the opposite, empty compartment and removing the diaphragm, the atom can be allowed to push the piston out, performing useful work equivalent to $kT \ln 2$.

Practically, predictability equates to a system's capability to utilize the correct energy extraction engine to harness energy from the memory unit. The architecture of this engine varies based on the informational value. Generally, the number of engines per memory unit should match the number of different information values that the unit can store. For a basic binary unit that stores either of the two binary numbers 0 or 1, only two engines are needed – a 0-engine and a 1-engine. These engines are designed such that the piston can be inserted from the right or left side of the cylinder, corresponding to the atom's location.

In this framework, the learning process can be conceptualized as comprising two integral sub-processes: 1) the passive learning process, which involves copying externally observed data patterns (such as sequences of 0s and 1s in the binary case) into the system's internal hardware, represented by appropriately ordered sequences of 0- and 1-engines; and 2) the active learning process, which entails the development of an automated dynamic mechanism. This mechanism applies the internally stored engine-sequences to the incoming external data sequences, ensuring real-time alignment between the two. The passive process is characterized by observation and recognition of external data patterns, while the active process focuses on reproducing these patterns through internal operations. If the theoretical aspect (passive learning) is executed correctly, the practical aspect (active learning) should yield immediate rewards in the form of additional energy. This energy can then be utilized for various needs of the learning system. Such a framework naturally aligns the learning objectives of a truly autonomous system with its survival prospects. The quality of learning, gauged by the accuracy of dynamic copying, directly influences the system's survival, measured in terms of the amount of consumable and usable energy obtained.

This learning process can be likened to a digital version of the well-established physical phenomenon of resonance, as detailed in our previous work [10], particularly in its relation to learning. Recall that the energy $E$ of a physical system can be increased when the movable part of the system, represented by a variable $x$ moves in the direction of the external force $F$ applied to it:

$$\frac{dE}{dt} = F\frac{dx}{dt} \tag{2}$$

If the external force $F$ varies over time, the system can only increase its internal energy by synchronizing the movement of its internal variable $x$ with external force $F$. This synchronization entails ensuring that the velocity of $x$ always matches the sign of $F$: positive forces correspond to positive velocity, and negative forces to negative velocity. In such scenarios, the right-hand side of Equation (2) will predominantly be positive, facilitating a steady increase in the system's energy. This coordination between the patterns of velocities and forces mirrors the alignment of external data patterns with internal action patterns, ensuring that the engines correspond appropriately with the incoming data.

While the external data sequence may be infinite, the sequences of internal engines are inherently finite. Consequently, the alignment between external data patterns (representing reality) and internal engine sequences (representing the system's models) can only be an approximation, achievable to varying degrees of accuracy and primarily on a statistical basis. This highlights the fundamentally statistical nature of the learning process and underscores the necessity of managing potential errors that may arise during energy extraction from incorrectly predicted data. In instances of erroneous predictions, the system risks losing energy instead of gaining it. To mitigate this, it is crucial to implement an error protection mechanism. Such a mechanism would optimize the rates of data extraction over the long term, taking into account the statistical likelihood of errors.

To illustrate how an error protection mechanism might function, consider a simple binary memory cell equipped with a specific energy extraction mechanism. This mechanism involves allowing the diaphragm, which divides the cylinder into two equal parts, to move from an initial position $L_1 = 1/2$ to a certain final position $L_2 = R$, where $R$ is a tunable characteristic of the system. If the atom is in the left compartment (representing the binary value 0), this movement will expand that compartment, generating useful work and thus energy equal to $E_0 = kT \ln 2R$. Conversely, if the atom is in the right compartment (representing the binary value 1) due to noise, the same movement will compress the compartment with the atom, leading to a negative energy growth of $kT \ln 2(1 - R)$. If the frequencies of the actual, 0, and noisy, 1, values are respectively $Q$ and $1 - Q$, then the average energy gain can be expressed as

$$E = kT(Q \ln 2R + (1 - Q) \ln 2(1 - R)) \tag{3}$$

It is evident that this energy is maximized when $R = Q$, yielding:

$$E(Q) = kT(Q \ln 2Q + (1 - Q) \ln 2(1 - Q)) \tag{4}$$

In a noise-free scenario, $Q = 1$, the energy extraction rate per bit reaches its maximum, aligning with the Landauer limit, $E = kT \ln 2$. However, in the case of maximal noise, $Q = 1/2$, the extractable energy drops to zero: $E = 0$. This phenomenon is an instance of the Kullback-Leibler divergence, which remains positive whenever $Q > 1/2$, a result of the logarithm function's convexity property.

It is important to note that optimal energy extraction is inherently constrained by errors leading to dissipation. This limitation is a natural consequence of the need for stabilization: excessively rapid learning is not only detrimental but also theoretically impossible, as it would contravene the second law of thermodynamics. The process of energy accumulation must, therefore, reach a point of stabilization. This phenomenon is also observable in resonant systems. Consider the modified version of equation (2):

$$\frac{dE}{dt} = F\frac{dx}{dt} - \gamma \left(\frac{dx}{dt}\right)^2 \qquad (5)$$

Here, the inclusion of a dissipative term, characterized by a friction coefficient $\gamma$, plays a crucial role in regulating energy growth, effectively stabilizing it when the right-hand side of equation (5) approaches zero. In this scenario, the fundamental concept of learning – aligning internal behavior (controllable) with external stimuli – remains unchanged. The velocity must still be proportional to the force, with a positive coefficient $\gamma^{-1}$ to achieve stabilization. However, this concept now acquires an additional quantitative dimension: the rate at which resources can be extracted from external data is limited by the nature of the data itself.

Summarizing the above, we can say that learning is a complex process of system's internal hardware optimization aimed at achieving maximal energy extraction rates. We can distinguish between the following two essential and complementary elements of this process: the optimization of the internal system organization (equivalent to the selection of the physical model whose energy is given by $E$ in the left-hand side of equation (5). Secondly, it involves minimizing unrecoverable energy losses due to dissipation, which corresponds to reducing the value of the friction coefficient $\gamma$ in the right-hand side of the same equation. Together, these elements constitute the core components of the learning process, each contributing to the system's overall efficiency and effectiveness in energy management.

## 3. On the Distinguished Role of Constants

The principle outlined previously leads to an important conclusion regarding the optimal operating conditions for the Autonomous Turing Machine (ATM). Specifically, the most advantageous scenario for the ATM occurs when the external data it encounters is constant. Considering the 1-bit memory model as an example, such data could be a sequence composed entirely of either 0s or 1s. In both instances, only a single, simple engine is required. This engine would consistently insert the piston up to the midpoint of the cylinder from the same side, thereby generating the maximum possible energy (equivalent to the Landauer limit) at each cycle. This energy can then be fully utilized to propel the ATM along the tape. Once processed or utilized, the information on the tape is effectively destroyed or randomized, increasing the entropy of the path left behind by the ATM. From this perspective, the ATM functions similarly to any conventional heat engine.

This schema is depicted in an illustrative diagram taken from paper [7]:

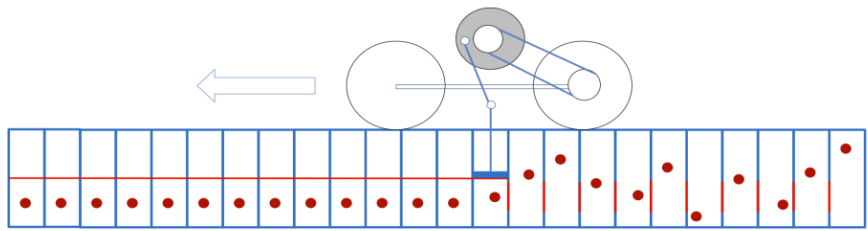

It demonstrates that in the scenario of constant data, the ATM operates in a straightforward manner, requiring only a single engine. However, the situation becomes more complex when the data is not constant but instead contains hidden regularities yet to be uncovered. Can the same engine, effective in extracting energy from constant data, be adapted for use with non-constant data by transforming the latter into a constant form before processing? The answer to this intriguing question is largely affirmative. This transformation process and its feasibility can be comprehensively understood within the abstract framework of the mathematical theory of dynamical systems.

Indeed, it is reasonable to assert that all data observed by an ATM is generated by some form of a dynamical system. Broadly speaking, a dynamical system can represent virtually anything that changes over time, encompassing everything around and within us. Whether it's processes in nature, society, or our own brains, all are dynamic in nature. To understand a system as a dynamical system is to comprehend how its past, present, and future states are interconnected. This understanding, or the ability to answer how these states relate, is essentially what we often refer to as intelligence.

It's fair to argue that every scientific field fundamentally revolves around the principles of dynamics. It begins with the discovery of these principles and culminates in their practical application. In this sense, there is no distinction between different scientific disciplines: the objective of Machine Learning aligns with that of Physics. The predictive models developed for various domains, such as finance, business, and social sciences, serve a similar purpose to the equations of Newton or Schrödinger in physics. They provide insights into future outcomes based on current knowledge. This parallel extends even to the mathematical structure of these models, which often share a common form.

This universality positions dynamical systems as an ideal framework for drawing connections between diverse and seemingly disparate fields. In this paper, our primary focus is on exploring the link between the informational and physical aspects of intelligence. The inherent system-independence of dynamical systems offers a promising avenue for understanding how intelligence might naturally arise in non-biological systems.

We can conceptualize learning as the process of uncovering invariants – combinations of data that remain constant or change very slowly. Once we identify such a stable combination, we label it as an object and integrate it into our knowledge network. This network is essentially a web of objects or invariants. The epistemological value of these invariants is significant; they are the elements worth memorizing. If they were to change constantly, recalling them when needed would be impractical and devoid of utility.

But how do we practically discover these invariants? It's known that any dynamical system can be described in infinitely many ways, each equally valid. Through simple coordinate transformations, we can navigate between these descriptions. The complexity of the system can vary depending on the chosen coordinates, leading us to question which coordinates render the system simplest, both theoretically and practically. When we first encounter a dynamical system, our choice of coordinates is often predetermined by the measuring instruments we use, defined by technical and historical factors. These initial variables, which we can directly measure, may not always be optimal for understanding the system. In their terms, the system might appear complex, with seemingly unpredictable changes over time. To grasp the nature of these changes and make them more predictable, our first step is to identify underlying regularities. In data science, these regularities are often referred to as patterns.

Why are patterns so captivating to us? Their allure lies in their ability to facilitate recognizability, which is a fundamental prerequisite for understanding. Recognizability hinges on the presence of elements that either remain constant or change in a predictable, repeatable manner. To assert that a phenomenon occurs regularly or forms a pattern is essentially to acknowledge the existence of an underlying, unchanging element. For instance, describing the regularity of sunrise and sunset is tantamount to recognizing the constancy of Earth's rotational speed. Thus, learning about new phenomena can be seen as a quest to uncover these unchanging components – islands of stability amidst the volatile ocean of data we encounter. Physicists approach this by seeking combinations of observable variables that are constant or change so slowly they can be practically utilized, often referring to these as 'conservation laws'. In the realm of dynamical systems, we term these 'dynamical invariants'. The more dynamical invariants we uncover, the deeper our understanding of the system becomes.

The challenge of identifying these invariants in a given dynamical process is intrinsically linked to finding a coordinate system where most variables remain constant. For example, when describing the Moon's orbit around the Earth, a polar coordinate system is optimal because it renders one coordinate (the radius) constant. Confronted with a new dynamical system, we are compelled to ask: How can we be sure these invariants exist? What is the maximum number of invariants we might discover? And how do we select the most effective coordinate system to maximize the utility of these invariants? These are the pivotal questions we aim to address in the subsequent sections.

## 4. Dynamical Invariants and Self Learning

Dynamical systems can be described through multiple equivalent mathematical frameworks. In the commonly used discrete version, the value of a vector variable $x(t) = x_1(t), \ldots, x_K(t)$ at time $t$ is defined in terms of its value at a previous time $t - \Delta$:

$$x(t) = F(x(t - \Delta)) \qquad (6)$$

Here $F(x)$ is a K-component vector function characterizing the system, assumed to be independent of time $t$, except through the time-dependent variable $x(t)$. We focus on the discrete version of dynamical equations, where the intervals $\Delta$ between successive observations are constant. To transition to the continuous version, often used in physics, we redefine $F$ as $F(x) = x + \Delta f(x)$ and consider the limit as $\Delta \to 0$. For a discussion of continuous dynamical invariants, refer to reference [16]. However, for consistency with computational language, we will continue with the finite $\Delta$ case.

It's important to note that system (6) remains covariant under any invertible coordinate transformation:

$$\xi = \Xi(x) \qquad (7)$$

This preserves the system's general form in the new coordinates:

$$\xi(t) = \Phi(\xi(t - \Delta)) \qquad (8)$$

The complexity of the new functions $\Phi$ may vary compared to the original functions $F$. A natural question arises: In which coordinate system is the form of these functions simplest, and what does this simplest form look like? To address this, consider that the most general solution of the system can be expressed as:

$$x(t) = X(c_1, \ldots, c_{K-1}, t + c_K), \quad k = 1, \ldots, K \qquad (9)$$

Here $c_i$, $i = 1, \ldots, K$ are arbitrary parameters, with the number $K$ determined by the initial conditions needed at $t = 0$ to fully define the future evolution of $x(t)$. These parameters can be viewed as functions of the initial positions. The inclusion of one of these parameters in combination with time in equation (9) stems from our assumption that the system's right-hand sides are time-independent, implying that shifting a solution in time yields another valid solution. Consequently, one of the constants must serve as a reference point for time.

Let us now our system as a set of $K + 1$ algebraic equations with $K + 1$ unknowns $t + c_0$ and $c_1, \ldots, c_{K-1}$. By solving these equations, we can express these unknowns in terms of the known coordinates $x$ leading to expressions like:

$$c_i = \Xi_i(x), \quad i = 1, \ldots, K - 1, \quad t + c_K = \Xi_K(x) \qquad (10)$$

These equations represent a specific invertible transformation to new coordinates $\xi_i$, $i = 1, \ldots, K$. This transformation is unique in that $K$ of the new variables are constants, and one variable behaves like time. Consequently, the new right-hand side functions $\Phi_i$ in this coordinate system are simplified to:

$$\Phi_i = \xi_i + \delta_{iK} \qquad (11)$$

This form is arguably the simplest possible for such functions. From this, we can infer that for any dynamical system of dimension $N + 1$ there exist $N$ functions of its dynamical variables that remain constant over time (these are our dynamical invariants). Additionally, there is one extra function among these variables that changes in the same manner as time does, barring an inconsequential constant.

This understanding leads us to two crucial insights:

1. Every dynamical system, regardless of its apparent complexity, is fundamentally simple at its core. Through an appropriate coordinate transformation, it can be reduced to a form where all its variables (except one) are constants, i.e., dynamical invariants. The sole remaining variable is simply time.
2. All dynamical systems of the same dimension are equivalent. For instance, if we have two systems, $S_1$ and $S_2$ both of dimension $N + 1$, they can be transformed into a trivial form $S_0$ using two invertible transforms $U_1$ and $U_2$. This implies that $S_1$ can be derived from $S_2$ through the combined invertible transform $U_1^{-1}U_2$.

These insights suggest that learning a new dynamical system involves finding a transformation that reduces its equations to this trivial form. While conceptually, mathematically, and philosophically profound, this approach may not be the most intuitive from the perspective of a learning agent. For such an agent, these high-level abstractions lack practical relevance. They do not inherently motivate the agent to embark on the discovery of dynamical invariants. This gap highlights that the impetus for learning cannot be fully explained by mathematics alone; additional considerations, particularly from physics, are necessary. One such consideration proposes a direct, energy-based reward mechanism for organisms engaged in the search for constants – the dynamical invariants. This perspective is particularly intriguing as it frames learning (viewed as an information solidification task) as a physical process aimed at energy accumulation. In the following section, we will explore this concept of constant-based learning from a physical standpoint.

This perspective elevates the concept of learning to the same level as fundamental survival activities like hunting or mining, where the objective is to search for food or other essential resources. Consequently, the process of data transformation, as described in section 1, which focuses on extracting dynamical invariants (or simply, constants) from data, becomes intrinsically linked to the process of converting these invariants into energy flows.

This concept is the foundation for our initial design of a universal learning machine based on physical principles:

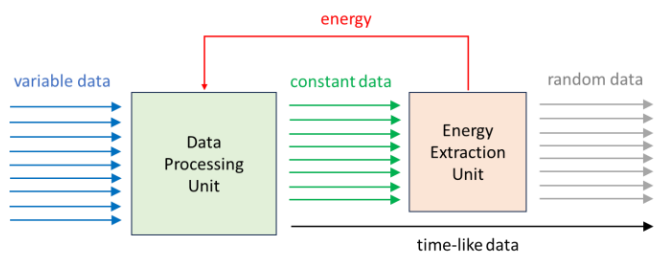

The depicted schema comprises two main components. The first is the data transform block, where input data is processed to produce an output with one time-varying component and multiple constant components. The second is the energy extraction block, which converts the constant data from the first block into random data. This randomization process, in turn, generates usable energy. The underlying idea is that this energy can be utilized for the internal needs of the system.

An enticing prospect for utilizing the generated energy is to power the data processing unit itself. If the combined system could be made energetically efficient – meaning the energy produced in the second block exceeds the energy required by the first – then we could regard it as a fully autonomous information-consuming system, capable of operating independently without external energy sources. This might seem miraculous, but such a scenario is feasible under a broad range of conditions. The key lies in the data processing block, which transforms the data of a K-dimensional dynamical system into its special degenerate form in terms of variables $\xi_i$. Remarkably, this block does not require any energy to function as it is entirely energy neutral. Why is this so? Because the transformation from the $x$- to $\xi$-variables is logically invertible. As demonstrated in [8,9], any logically reversible transformation can be executed in a thermodynamically reversible manner, meaning it doesn't necessitate additional energy consumption[2]. This realization opens up the possibility of constructing a physical device for any finite-dimensional dynamical system. Such a device, comprising a reversible data-processing unit and an energy extraction block, could convert incoming data flows into energy.

---

[2] Consider the NOT-gate in our cylinder model as a simple example. To change the value from 0 to 1, or vice versa, one only needs to rotate the cylinder 180 degrees, effectively swapping its left and right parts. In more complex scenarios, all computations can be rendered energy-neutral using reversible gates, such as Fredkin or Toffoli gates. The principle behind these gates is to retain 'garbage bits' instead of discarding them. However, it's important to note that logical reversibility implies thermodynamic reversibility (and thus energy neutrality) only in the context of adiabatically slow processes. Computation, by its nature, involves change, which is typically accompanied by energy dissipation. To minimize dissipation, the speed of the processes must be reduced, ideally to zero. For the purposes of our discussion, we will not consider dissipation-driven energy effects, bearing in mind that their impact can be controlled by moderating the speed of internal changes.

Does this mean we can disconnect the energy supply from the Data Processing block and let it operate independently? Not quite. The challenge lies in the fact that each type of incoming data necessitates its own specific reversible transformation. While the transformation itself, once established, does not consume energy – thereby ensuring a net gain in energy production with the correct transform – the initial process of identifying the appropriate transformation is not inherently energy-neutral. Energy expenditure is required to determine the right or optimal transformation. Similarly, the energy extraction block isn't automatically optimized for every type of incoming data; it requires separate fine-tuning for each data variant.

The encouraging aspect, however, is the substantial potential reward once such a transformation is discovered. Since maintaining the operation of the correctly identified transform doesn't require additional energy, all the energy produced by the second block can be utilized in numerous ways.

Recalling our earlier definition of learning as the process of transforming a flow of changing data into a flow of constant data, we can now assert with confidence that learning demands an energy investment. Yet, if this process is successfully completed, it can unlock access to seemingly unlimited resources. This isn't just a metaphorical statement; it's grounded in a profound physical reality.

## The Universal Self-Propelled Learning Architecture

As previously mentioned, in an ideal scenario where data is noise-free and strictly adheres to the dynamical equations of a finite dimension $K$, the output of the reversible transform comprises $K-1$ constants. The last component, however, is not constant – it is a time-like variable that continuously increases. This variable's role is crucial for maintaining the reversibility of the transformation: different $t$ values act as 'public keys' enabling the decoding of data encoded by the algorithm $F$. Yet, due to its constant change, this time-like variable is not directly suitable for energy generation. Nevertheless, there is a straightforward (albeit indirect) method to convert this time variable into a constant. This involves subtracting the delayed time-like signal, $t - \Delta$, from the original one, $t$ in a reversible manner. To ensure reversibility, the input signal $t$ is preserved in the output and then fed through a delay latch $\Delta$ creating another input, $t - \Delta$, for the subtraction block. The result of this subtraction is a constant $\Delta$ – effectively adding an additional $K$-th constant to the existing $K-1$ constants. constants. This operation balances the number of variable inputs with constant outputs.

These $K$ constants can now be directed to the energy extraction block. It's important to note that the internal organization of this block might vary depending on the specific set of constants it receives. Each constant can be visualized as a series of 1-bit registers composed of symmetric cylinders, as described earlier. Conceptually, the process of extracting energy from these registers remains consistent across all registers, with the only variation being the side (left or right) from which the piston is inserted, determined by the atom's position in the cylinder (i.e., the binary value it represents). This makes the energy extraction engine dependent on the constants. Should there be a sudden change in the set of constants – perhaps due to alterations in the input data or its transformation – the energy extraction engine's efficiency could significantly decrease, potentially even becoming negative.

To avert situations where the efficiency of the energy extraction engine drops, we can employ a similar strategy to the one previously described – by adding an additional $K$ Reversible Transformation (RT) units, each equipped with its corresponding delay latch. In this setup, performing an XOR operation between a constant and its delayed value will invariably result in zero, as constants are inherently delay-invariant. It's important to note that the objective here is not merely to obtain a zero value (which, in terms of energy extraction, is no more advantageous than any other constant). Rather, the key advantage lies in standardizing the output for any input constant. This standardization enables the design of identical energy extraction engines, all sharing the same architecture regardless of the input constant's value. This uniformity not only facilitates the conservation of hardware resources but also provides a significant benefit: it safeguards the energy production rates against abrupt changes in the sets of constants.

An alternative approach, which is more universal and conceptually simpler, can also be employed within the RT unit. This method involves implementing a transformation that reflects the original form of the system's memoryless dynamical equations (). Specifically, the transformation is executed as follows:

$$z(t) = x(t) - F(x(t - \Delta)) \tag{12}$$

This operation is carried out in a reversible manner. The subsequent goal is to iteratively refine the function $F$ such that the output $z(t)$ becomes as close to zero as possible. This approach essentially involves reconstructing the form of $F$ to minimize the output, aligning it with the desired state. The general schema for this process is illustrated below:

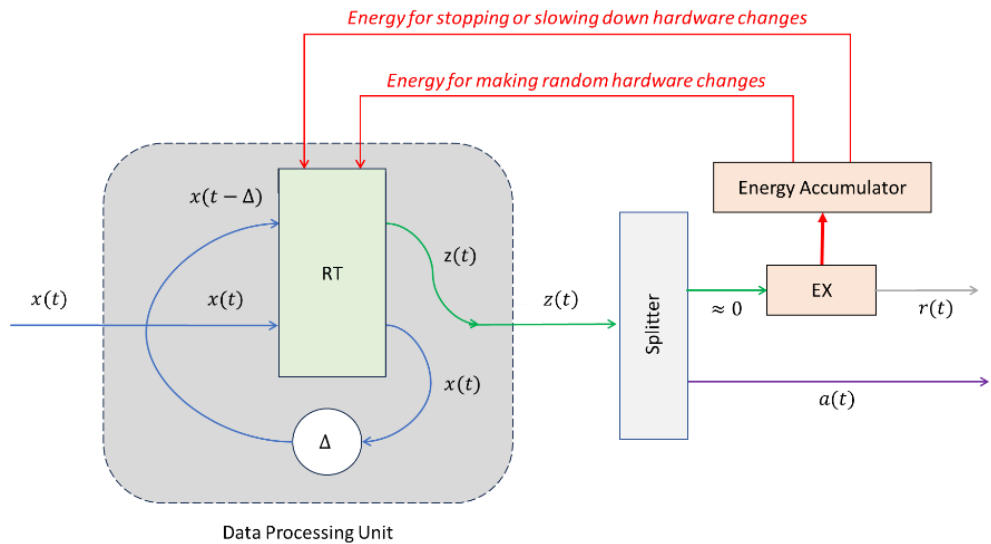

As observed, the memoryless form necessitates a set of delay latches with a unit depth equal to $\Delta$. This is expected since we are dealing with a system of equations of motion that connects the present moment with the previous one. The resultant signal $z(t)$ should ideally be as close to zero as possible. However, achieving perfect 'zero-ness' is impractical; we can only aim for its statistical proximity to zero. This means acknowledging the presence of non-zero noise within its components. In cases where the data is imperfect, the $z(t)$ signal can be divided into two sub-flows. The first sub-flow, depicted by the green line, includes components nearly zero, possibly with a negligible amount of noise manageable by a zero-signal-focused energy extraction engine. The second sub-flow, represented by the purple line, comprises rapidly fluctuating components significantly deviating from zero. These appear random to the AI agent but are still influenced by the input in ways not fully comprehended by the agent. This input-dependence renders them valuable, as their externalization could potentially alter data inputs and lead to the emergence of new, less noisy zero components.

The energy extracted from the zero data was earlier characterized as useful energy – that is, energy the agent can utilize for its own needs. But what needs might an agent have? Approaching this question pragmatically, and drawing inspiration from common sense, the optimal use of this energy appears to be reinvestment in further learning. This approach is pragmatic because it directly leads to new potential energy sources. However, it also creates a closed logical loop: we learn to accumulate resources, and we accumulate resources to learn. This seemingly endless cycle is what we often perceive as 'life', or at least the part of it we are conscious of. Learning necessitates modifications in the hardware architecture of the RT module. But what is the most universal mechanism for implementing these changes?

One practical solution for enhancing the learning algorithm involves dividing the energy flow exiting the Energy Extraction (EX) unit into two distinct parts using a specialized two-channel splitter. One channel of this splitter maintains a constant bandwidth, while the bandwidth of the other channel varies in proportion to the amount of energy extractable from the $z$ signal. This extractable energy amount is indicative of the quality of learning, or equivalently, the agent's overall intelligence level. The concept here is to allocate the first, steady stream of energy to gradually effectuate changes in the hardware architecture of the RT-block, such as the slow reconfiguration of its reversible gates. Concurrently, the second, variable energy stream acts as a control mechanism, capable of decelerating or even halting these hardware modifications. This setup ensures that the quest for the most energy-efficient architecture persists until an optimal configuration is achieved, at which point the control mechanism activates, and the search ceases.

While this approach may not guarantee the most effective learning algorithm, it possesses a crucial feature of adaptability or the capacity to relearn in response to changes in external data. Alterations in the external data patterns signify shifts in the dynamical model, disrupting the conversion of original data into the 'zero' form. Such disruptions result in a decrease in energy generation, which in turn deactivates the control mechanism. Consequently, the hardware meta-parameters (the search mechanism) are reactivated, initiating a new cycle of optimization in response to the changed data environment.

This approach can be extended to time-series forms of dynamical equations, where the current value of a dynamical variable $x(t)$ is determined by its $N$ equidistantly spaced past values:

$$x(t) = G(x(t - \Delta), \ldots, x(t - N\Delta)) \qquad (13)$$

It's important to note that equations (13) and (6) are equivalent and can be converted into one another. This is achieved by treating the past values of the variable $x$ as new independent components of the vector $\vec{x}$ and vice versa. To generalize this concept, we utilize $K$ delay latches and a $K \times K$ reversible transform, as depicted in the accompanying picture:

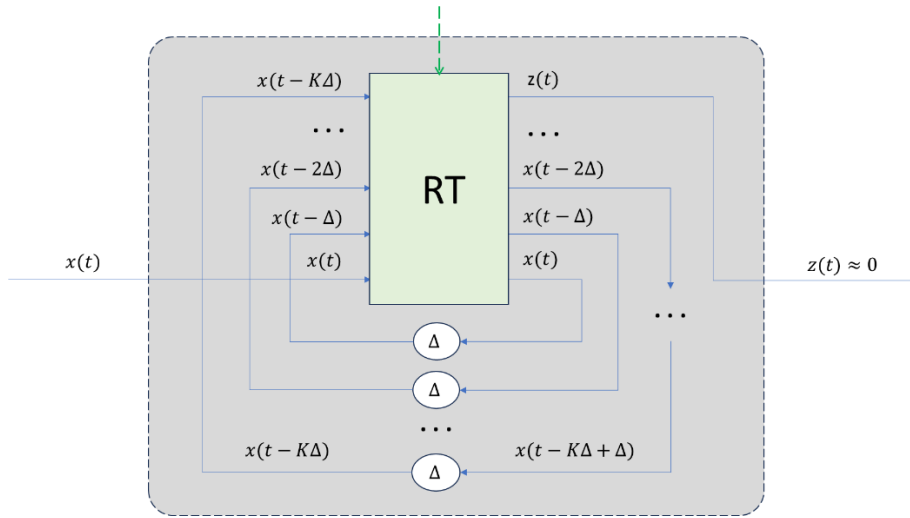

In this illustration, the delay latches need not be physically located within the unit. They can also be conceptualized as markers that signify the agent's position along its trajectory through the environment.

## 5. The Caterpillar Model and Memory Externalization

Just as humans utilize external tools, ranging from simple paper notes to sophisticated computing systems, to augment our brain's computational capabilities, AI agents can similarly benefit from leveraging their external environment as an extension of their computational resources. The immediate surroundings of an agent's 'body' present an ideal opportunity for such an extension, particularly for short-term memory units like the delay latches discussed earlier. These surroundings are conveniently accessible for brief periods while the agent is in close proximity. A prime example of this concept of hardware resource externalization is what we will refer to as the 'Caterpillar Model' of the universal learning architecture.

The Caterpillar Model, comprising the same four essential blocks as depicted in the previous picture and thus identical to the general model, operates through the following seven steps, each illustrated pictorially:

1) Positioning: The caterpillar positions its Reversible Transformation (RT) block over the $n - m + 1$ cells of the tape, which store the numbers $x_n = x(t), \ldots, x_m = x(t - K\Delta)$:

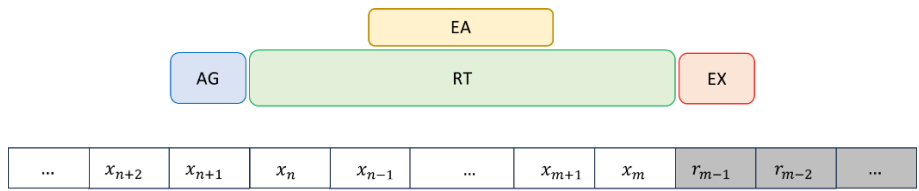

2) Reading and Processing: The RT block reads the content of these cells into its internal input registers and begins processing them in a reversible manner.

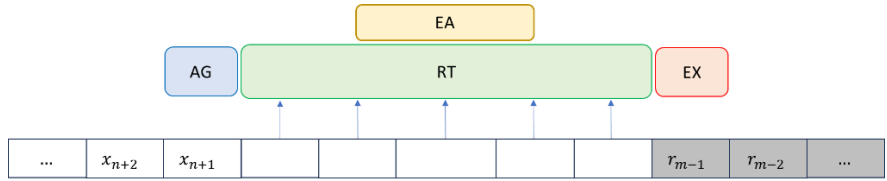

3) **Writing Output:** After processing, the RT block writes the output back to the tape. The first $n - m$ numbers of this output match the original numbers $x_n, \ldots, x_{n-m+1}$ previously on the tape, with the last number being 0:

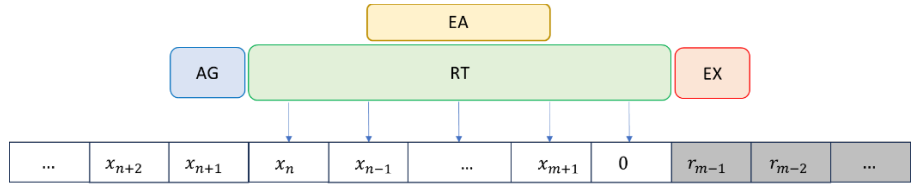

4) **Movement:** The Action Generation (AG) block utilizes energy from the Energy Accumulation (EA) unit to move the caterpillar to the next position:

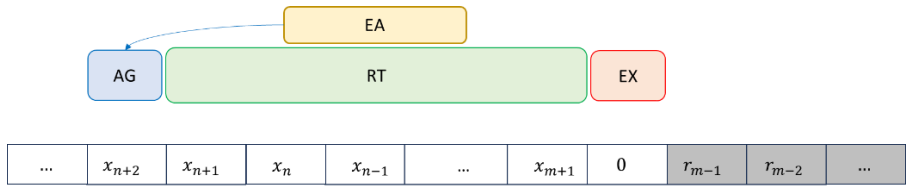

5) **Absorption:** The Energy Extraction (EX) block absorbs the zero value, preparing it for energy extraction:

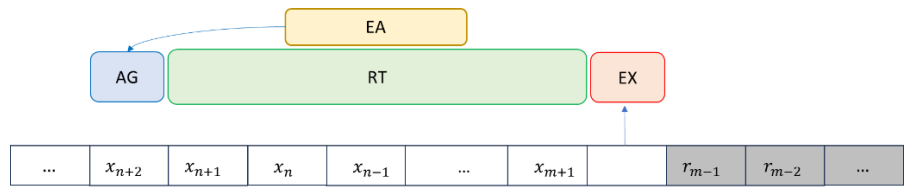

6) **Energy Extraction:** The EX block extracts energy from the zero signal and channels it to the energy accumulator, offsetting the energy used in the previous step. The byproduct of this process – random data – is written back to the tape in place of the zero signal:

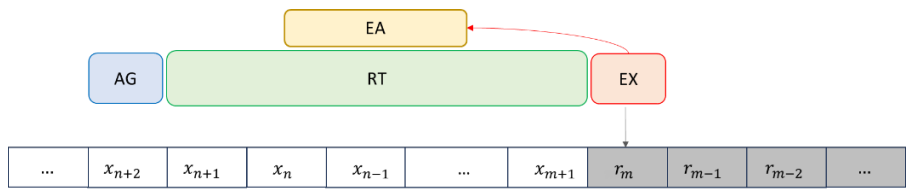

7) **Reset and Shift:** The caterpillar returns to its original state as in Step 1 but shifted forward by one cell. The RT block now covers the $n - m + 1$ cells of the tape, containing the updated numbers $x_{n+1} = x(t + \Delta), \ldots, x_{m+1} = x(t - K\Delta + \Delta)$:

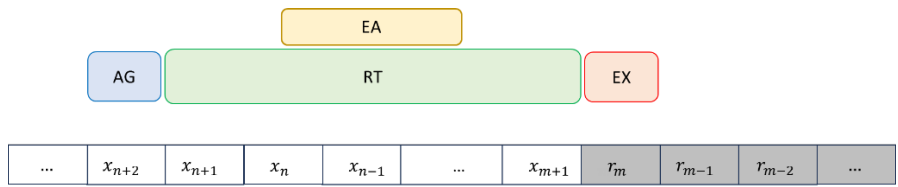

This sequence completes a single cycle of the Caterpillar Model, after which the process can be repeated indefinitely.

One of the key attributes of the caterpillar model is its elimination of time as a global synchronizing factor. In models relying solely on internal latches, the necessity for uniform delay durations and synchronization with the rate of new data arrival is ambiguous. However, in the caterpillar model, this issue is resolved as the caterpillar autonomously decides when to acquire new data and allocates its processing time accordingly. Here, time transforms from a global variable into a local one. At first glance, it seems that the temporal variable $t$ is replaced by the ordinal number $n$ representing the step count. Yet, the actual extension of this count becomes irrelevant. What gains practical significance is the accurate counting of cell numbers within the caterpillar's body, specifically within the range of $n - m$. Beyond this range, the concept of the 'next cell' is the only relevant factor, allowing for a flexible tape topology and facilitating extension to multi-dimensional scenarios.

In a multi-dimensional environment, the caterpillar gains additional degrees of freedom, enabling it to choose and control its movement direction. This capability provides a straightforward and efficient means of altering the character of the data it consumes. For instance, consider an environment structured as a memory-field on a 2D rectangular lattice. Each lattice cell, or individual memory unit, is defined by an integer vector $\vec{n}$ (its address) and a stored value $x_{\vec{n}}$. The caterpillar forms a certain linear shape in the space which may have a complex form rather than be straight. The caterpillar occupies a linear shape within this space, which can vary in complexity:

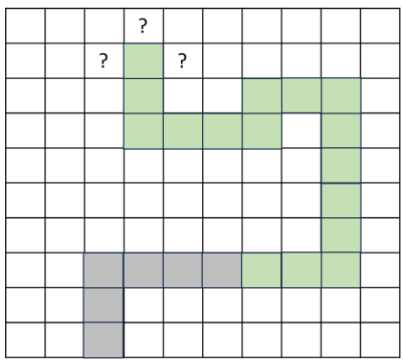

The accompanying picture illustrates a caterpillar occupying 16 green cells in a 2D space, leaving a gray trail of processed cells. At each step, it has three directional options. The decision-making regarding these turns is a responsibility of the Action Generation (AG) block.

Another intriguing aspect of the caterpillar model is that not only the input and output of the Reversible Transformation (RT) block but also the entire computation process can be externalized from the caterpillar body. To demonstrate this, consider the general reversible transformation of three input bits into three output bits. Such a transformation can typically be executed using a series of elementary reversible gates, like Fredkin or Toffoli gates. Examining the physical realizations of these gates reveals that no additional intermediate memory units are necessary for these calculations – only specific hardware manipulations are required. All these hardware manipulations can be energy-neutral, provided they are performed slowly enough to minimize energy dissipation.

In summary, the RT block theoretically does not require any internal registers, and all computations can be conducted solely through hardware manipulation mechanisms. This principle also applies to the Energy Extraction (EX) block. In light of the aforementioned characteristics of the RT block, the operations of the EX block can be carried out directly on the tape. This is achieved by inserting the piston into a cell containing a zero value and allowing it to generate useful work after the diaphragm is removed.

## 6. Conclusion

There is a fascinating hypothesis suggesting that intelligence might spontaneously emerge in certain low-entropy environments as a stable fluctuation. This emergence is contingent upon two conditions: 1) the stability of the fluctuation is sustained by energy extracted from the environment, and 2) the fluctuation is capable of resonating with external forces acting upon it. The rationale behind these conditions leading to phenomena identifiable as intelligence is as follows. Resonance with an external force implies the existence of internal mechanisms that can replicate the pattern of these forces, essentially predicting their behavior. Moreover, a system entering a resonant state with its environment effectively harnesses energy from it. These observations forge a connection between a) the ability to make predictions, b) understanding the mechanisms of external processes, and c) the process of energy accumulation. In essence, this links intelligent behavior (characterized by properties a and b) to the pragmatic need for survival, thereby also explaining the motivation for continuous learning and knowledge expansion.

The digital counterpart of this concept can also be articulated, particularly by leveraging the concept of information as a physical state of low entropy. Utilizing the principles of thermodynamic information theory, it becomes evident that correctly predicted information can serve as a tangible energy resource. This is 'real' in the sense that it is feasible to construct a physical device that interacts with a memory cell, where information is stored, in such a manner that allows for the immediate conversion of this information into useful work or energy storage for later use. A remarkable feature of any autonomous system, whose existence or well-being hinges on its energy reserves and which is equipped with such a device, is an inherent drive to learn. This is because learning enhances predictive capabilities, which in turn grants access to energy resources. All facets of this process can be precisely quantified and formalized, paving the way for the conceptualization of hypothetical models known as Autonomous Turing Machines (ATMs). These devices are envisioned to navigate through external environments (represented by N-dimensional generalizations of tapes used in standard Turing Machines), seeking patterns whose 'consumption' is vital for their survival.

It can be shown that the above physical-resonance-based and information-theory-based versions are equivalent because both are based on the process of matching the externally observed patterns by internally built patterns. In the physical case these internal patterns are formed by system's velocities, while in the digital case they are formed by the sequences of energy-extracting engines. Both velocities and engines symbolize actions of these systems.

In this paper we tried to better understand the possible meta-architectures of these digital systems and the role played in them by actions, through which these systems can actually evolve. It was shown that this architecture allows a clear modular form with three well distinguished units specialized in a) energy extraction from a 'noisy zero' signal, b) data processing unit consisting of energy-neutral reversible computation blocks and a number of delay latches together responsible for generation of a 'noisy zero' and d) the control system fed by the energy extracted from the 'noisy zero' and responsible for rewiring the hardware elements of the data processing block in such a way that to reduce the amount of noise in 'noisy zero' signal and, through this reduction, achieving the maximally possible energy production rates.

The details on how the controller could work is still not quite clear to us, which partially is caused by the fact that we use the mixture of diverse methods and paradigms instead of working within a single and well-formalized theory. We firmly hold the conviction that every facet of intelligence can, in time, be comprehensively delineated and grasped solely through the lens of physics, encompassing both its classical and quantum dimensions. Nevertheless, we believe that the language employed in this paper, intertwining elements of information theory, dynamical systems, and thermodynamics, can temporarily remain a useful approach for conceptualizing intelligence as a research domain. This approach notably underscores its inherently interdisciplinary nature at present. A primary motive for integrating these somewhat loosely related fields is to tap into the vast expanse of knowledge that has been independently amassed within each discipline.

While employing our chosen language framework, it's crucial to acknowledge its formal limitations and the possibility of internal contradictions leading to logical inconsistencies. We have endeavored to minimize these inconsistencies, yet certain aspects, particularly regarding the treatment of actions within this framework, remain ambiguous. Actions are pivotal in the learning process, representing change, and from an information theory perspective, this implies the generation or modification of information. The information pertaining to actions is similar to observational data, requiring pre-existing memory units for their creation, modification, storage, and reuse. However, in our discussions about changes in what could be termed 'hardware' components, we deliberately refrained from interpreting these within a thermodynamic information framework. Instead, we opted for descriptions external to this model, utilizing alternative frameworks like solid-state physics. For instance, when discussing the stopping position of a piston in a cylinder, we did not delve into the memory requirements for storing and using this information; we simply acknowledged the piston's position.

This approach mirrors the relationship between quantum mechanics and classical mechanics, where the former views the latter as a special case but still relies on it for foundational concepts. Similarly, the thermodynamics of information theory calls for an in-depth examination of concepts that, while formally outside its scope, are inherently linked to it in a broader sense. This includes elements of an intelligent agent's internal architecture (its 'skeleton'), which we have referred to as 'hardware'. These components fulfill two roles: they delineate the boundaries of thermodynamic models of memory units and computational devices, and they are crucial for managing computational processes and energy extraction at the thermodynamic or 'software' level. Such management is essential for learning, as it involves algorithmic changes, necessitating adjustments in wiring topology, which in turn depends on an understanding of the dynamics of the hardware elements. Therefore, to fully articulate the learning process within our model, we must incorporate principles from mechanics or solid-state physics. This integration inevitably makes the initial model more complex and less transparent, posing challenges for further advancements in the chosen direction.

Although the challenges in our model appear to stem from its foundation on a diverse blend of models and paradigms, the actual issue is more profound and lies in the prevailing understanding of computational models, especially in their application to the study of learning as a natural evolution of intelligence. In contemporary AI research, 'computational model' is often synonymous with 'software model.' This perspective emphasizes software and the modifications necessary to enhance intelligence, assuming that the hardware is sufficiently versatile to support these changes. Modern AI systems are typically designed with highly universal hardware architectures, capable of handling a wide range of computations and data types. However, this universality underscores the limitations of current computational models in addressing the natural evolution and learning processes of intelligence. In self-driven and self-evolving intelligent systems, the paradigm shifts: it is the hardware changes, rather than software updates, that should primarily drive the development of intelligence. Even minor software adjustments, like tweaking parameters, presuppose a degree of hardware flexibility, a condition often taken for granted. Yet, for naturally evolving systems, the spontaneous generation of additional hardware resources (such as memory space or processing units) for potential future needs is highly unlikely.

When comparing the explanatory power of purely physical versus digital models for the emergence and evolution of intelligence, it appears that physical models may have a better chance of prevailing. However, to do so, they must also evolve to become less restrictive than the models typically employed for describing well-defined classes of physical phenomena.

In this discussion, the constant and zero signals, which have been central to our approach, warrant further reflection. We believe that these concepts are crucial for understanding key aspects of intelligence, not just from a digital perspective, as presented in this paper, but also in physical models. Remarkably, the importance of constants is also evident in physical models of resonance, particularly when expressed in action-angle coordinates. Here, periodic or quasi-periodic external forces can be represented as constants coupled with constant internal velocities. This suggests that pursuing research in purely physical models might reinforce the idea that constants and data invariants are integral to understanding intelligence.

The process of learning through the discovery of data invariants offers several epistemological advantages, notably simplicity. Intelligence could be characterized as the quest to find a distinguished coordinate system where the universe appears at its simplest. Unlike patterns, which are inherently diverse and complex, constants offer a more economical description. They can be uniquely identified by a single number – their value and duration, representing minimal algorithmic length. The zero signal is even more efficient, requiring only its duration for description. This efficiency makes constant signals particularly memory-friendly, as they require fewer resources to memorize. Moreover, constants are the only form of information that truly merits memorization, as rapidly changing data lacks practical value for recall. Recognizing and adapting to constant features is faster, as they are, by definition, unchanging over time. For instance, understanding the dynamical invariant of an oscillation allows for quicker identification than waiting for a complete oscillation period.

These considerations are implementation-independent, suggesting that the ideas discussed in this paper could significantly influence future research. By focusing on the role of constants and zero signals in various models of intelligence, we open new avenues for exploring how intelligence can be understood, modeled, and potentially replicated.

# References

[1] J. C. Maxwell, The theory of heat (Appleton, London, 1871).

[2] L. Szilard, "Uber die entropieverminderung in einem thermodynamischen system bei eingriffen intelligenter wesen", Z. Phys. 53, 840 (1929).

[3] R. Landauer, "Irreversibility and heat generation in the computing process", IBM J. Res. Dev. 5, 183 (1961).

[4] Charles H. Bennett, "Thermodynamics of Computation -- a Review", International Journal of Theoretical Physics, Vol 21, No 12, 1982

[5] Feynman, R. P. & Hey, A., 2000. "Feynman Lectures On Computation". Westview Press; Revised ed.

[6] Plenio, M. B. & Vitelli, V., 2001. The physics of forgetting: Landauer's erasure principle and information theory. Contemporary Physics, 42(1), pp. 25-60.

[7] Alex Ushveridze, "Can Turing machine be curious about its Turing test results? Three informal lectures on physics of intelligence", arXiv:1606.08109, (2016)

[8] Fredkin, E. & Toffoli, T., 1982. "Conservative Logic". International Journal of Theoretical Physics, 21 (3-4), p. 219–253.

[9] Perumalla, K., 2014. "Introduction to Reversible Computing". Chapman & Hall/CRC Computational Science.

[10] Alex Ushveridze, "On Physical Origins of Learning", arXiv:2310.02375 [q-bio.NC], 2023

[11] Zhongkai Hao, Songming Liu, Yichi Zhang, Chengyang Ying, Yao Feng, Hang Su, Jun Zhu, "Physics Informed Machine Learning: A Survey on Problems, Methods and Applications", arXiv:2211.08064v2 [cs.LG] 2023

[12] G. E. Karniadakis, I. G. Kevrekidis, L. Lu, P. Perdikaris, S. Wang, and L. Yang, "Physics-informed machine learning," Nature Reviews Physics, vol. 3, no. 6, pp. 422–440, 2021.

[13] Ferran Alet et al. "Noether Networks: Meta-Learning Useful Conserved Quantities". In: ArXiv abs/2112.03321 (2021).

[14] Seungwoong Ha, Hawoong Jeong "Discovering conservation laws from trajectories via machine learning", arXiv:2102.04008 [cs.LG] 2021

[15] Jason A. Platt, Stephen G. Penny, Timothy A. Smith, Tse-Chun Chen, Henry D. I. Abarbanel, Constraining Chaos: Enforcing dynamical invariants in the training of recurrent neural networks, arXiv:2304.12865 [cs.LG], 2023

[16] Alex Ushveridze, "Classical Lagrange formalism for non-conservative dynamical systems", arXiv:2212.12409, (2022)